\documentclass[lettersize,journal]{IEEEtran}
\usepackage{amsmath,amsfonts}
\usepackage{algorithmic}
\usepackage{algorithm}
\usepackage{array}
\usepackage[caption=false,font=normalsize,labelfont=sf,textfont=sf]{subfig}
\usepackage{textcomp}
\usepackage{stfloats}
\usepackage{url}
\usepackage{verbatim}
\usepackage{graphicx}
\usepackage{cite}
\usepackage{booktabs}
\usepackage{algorithm}
\usepackage{amsmath}
\usepackage{amssymb}
\usepackage{mathtools}
\usepackage{amsthm}
\usepackage{multirow}
\begin{document}

\title{MEC-Quant: Maximum Entropy Coding for Extremely Low Bit Quantization-Aware Training}

\author{
Junbiao Pang, Tianyang Cai, Baochang Zhang

\IEEEcompsocitemizethanks{
\IEEEcompsocthanksitem J. Pang, T. Cai and are with the Faculty of Information Technology, Beijing University of Technology, Beijing 100124, China (e-mail: \mbox{junbiao\_pang@bjut.edu.cn}).

\IEEEcompsocthanksitem  B. Zhang is with the University of Chinese Academy of Sciences, Chinese Academy of Sciences (CAS), Beijing
100049, China, and the Institute of Computing Technology, CAS, Beijing
100190, China (email: bczhang@139.com).
 }
}

\maketitle

\begin{abstract}

Quantization-Aware Training (QAT) has driven much attention to produce efficient neural networks. Current QAT still obtains inferior performances compared with the Full Precision (FP) counterpart. In this work, we argue that quantization inevitably introduce biases into the learned representation, especially under the extremely low-bit setting. To cope with this issue, we propose Maximum Entropy Coding Quantization (MEC-Quant), a more principled objective that explicitly optimizes on the structure of the representation, so that the learned representation is less biased and thus generalizes better to unseen in-distribution samples. To make the objective end-to-end trainable, we propose to leverage the minimal coding length in lossy data coding as a computationally tractable surrogate for the entropy, and further derive a scalable reformulation of the objective based on Mixture Of Experts (MOE) that not only allows fast computation but also handles the long-tailed distribution for weights or activation values. Extensive experiments on various tasks on computer vision tasks prove its superiority. With MEC-Qaunt, the limit of QAT is pushed to the x-bit activation for the first time and the accuracy of MEC-Quant is comparable to or even surpass the FP counterpart. Without bells and whistles, MEC-Qaunt establishes a new state of the art for QAT. Our code is available at this https URL and has been integrated into MQBench (this https URL)
\end{abstract}

\begin{IEEEkeywords}
Quantization, Maximum Entropy Principle, Long-tailed Distribution, Mixed of Expert
\end{IEEEkeywords}

\section{Introduction}
\IEEEPARstart{W}{ith} the rapid development of deep learning and the rapid growth of massive data, artificial intelligence has concentrated on the two poles of edge computing model and cloud model. However, the high training and storage costs have hindered the production and deployment of deep learning models, and model compression has therefore attracted great attention. Model compression techniques are further divided into neural architecture search~\cite{zoph2016neural}, network pruning~\cite{han2015deep}, and quantization~\cite{jacob2018quantization, xu2023q}. Among these methods, quantization plays an important role in reducing memory consumption and accelerating inference due to low bit weights and activations.

According to the different optimization objects, quantization can be divided into Post-Training Quantization (PTQ) ~\cite{nagel2020up,li2021brecq,wei2022qdrop} and  Quantization-Aware Training (QAT)~\cite{esser2019learned,nagel2022overcoming}. PTQ only uses a calibration set composed of a small portion of the training set for fine-tuning, without the need to retrain the entire training set. Therefore, PTQ is very concise and efficient, but it also faces serious accuracy loss at extremely low bits. QAT uses the entire training set for retrain, simulating errors caused by quantization during inference while optimizing weights and quantization parameters. It still maintains good performance at extremely low bit. Therefore, at extremely low bit, QAT is widely used in both academia and industry.

However, current QAT still obtains inferior performances compared with the Full Precision (FP) counterpart. In this work, we argue that quantization inevitably introduce biases into the learned representation, especially under the extremely low-bit setting. In this work, we believe that this is because of the quantization operation itself causing bias in the learned representation, leading to performance degradation. This deviation will accumulate in the backbone, resulting in cumulative errors that are reflected in downstream tasks. To solve this problem, we propose Maximum Entropy Coding Quantization (MEC-Quant), a more principled objective that explicitly optimizes on the structure of the representation, so that the learned representation is less biased and thus generalizes better to unseen in-distribution samples.

However, due to the fact that features in deep learning are high-dimensional vectors, it is very difficult to directly calculate feature entropy and the computational complexity is relatively high. To solve this problem, we propose to leverage the minimal coding length in lossy data coding as a computationally tractable surrogate for the entropy~\cite{cover1999elements}. The log-determinant term costs the most computation in the coding length function. By using Taylor expansion on the matrix, we further accelerate and estimate it. As shown in Fig, high dimensional features usually have the characteristic of long-tailed distribution, and Taylor expansion at a certain point alone cannot meet the accuracy of estimation. To address this issue, we propose using Mixture Of Experts (MOE) for multi-point expansion, which not only ensures estimation accuracy but also speeds up operations. Our contributions are as follow:

\begin{itemize}
\item We believe that the quantization operation itself during the quantization process can cause bias in the learned representation, leading to performance degradation. We propose Maximum Entropy Coding Quantization (MEC-Quant), a more principled objective that explicitly optimizes on the structure of the representation, so that the learned representation is less biased and thus generalizes better to unseen in-distribution samples.
\item To accelerate the calculation and accuracy of feature entropy, we propose to leverage the minimal coding length in lossy data coding as a computationally tractable surrogate for the entropy. To address the issue of inaccurate estimation caused by the long tail distribution phenomenon in high-dimensional features, we derive a scalable reformulation of the objective based on MOE that not only allows fast computation but also handles the long-tailed distribution for weights or activation values.
\item We propose a training paradigm for QAT based on MEC named MEC-Qaunt,.The proposed approach, a simple, novel, yet powerful method, is easily adapted to different neural networks. MEC-Qaunt pushes the performances of QAT models towards, and even surpasses that of FP32 counterparts. Extensive experiments on different models prove that our method set up a new State-Of-The-Art (SOTA) method for QAT.
\end{itemize}

\begin{figure*}[t!]
    \centering
    \includegraphics[width=1.0\linewidth]{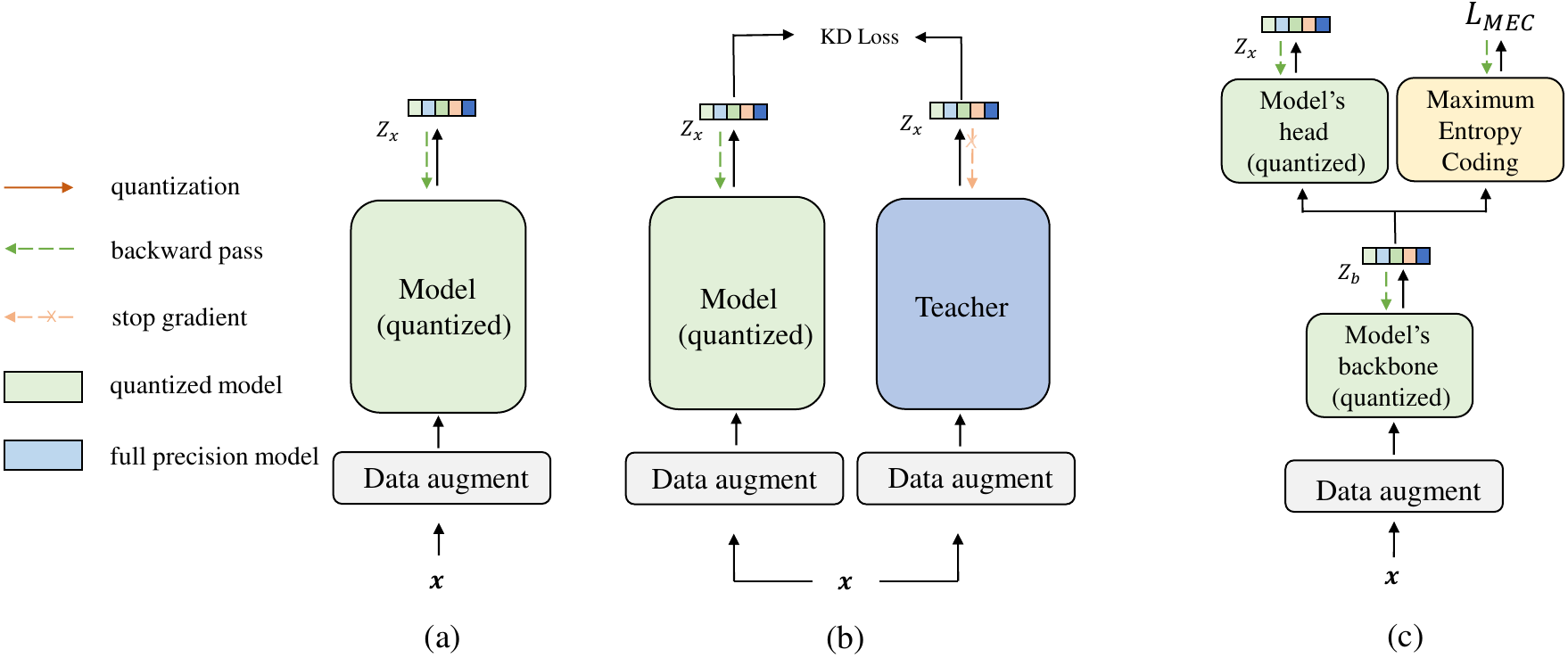}
    \caption{Comparison between vanilla QAT, quantization with KD~\cite{polino2018model} and our method. (a) the overview of vanilla QAT. (b) the overview of quantization with KD. (c) the overview of our method.}
    \label{fig:qat-kd-mec-comparision}
\end{figure*}

\section{Related Work}
\textbf{Quantization-Aware Training.} The idea of QAT is to minimize quantization errors by the complete training dataset, enabling the network to consider task-related loss. There are two research directions in QAT: 1) one primarily focuses on addressing quantization-specific issues, such as gradient backpropagation caused by rounding operations~\cite{bengio2013estimating,lee2021network} and weight oscillations during the training stage~\cite{nagel2022overcoming}; 2) the other integrates quantization with other training paradigms.

For the first category, STE~\cite{bengio2013estimating} employed the expected probability of stochastic quantization as the derivative value in backpropagation. STE assumes the derivative of the rounding operation is defaulted to 1, failing to accurately reflect the actual quantization error.
EWGS~\cite{lee2021network} adaptively scaled quantized gradients based on quantization error, compensating for the gradient. PSG~\cite{kim2020position} scaled gradients based on the position of the weight vector, essentially providing gradient compensation. {DiffQ}~\cite{defossez2021differentiable} found that STE can cause weight oscillations during training and thus employed additive Gaussian noise to simulate quantization noise. It is overcoming Oscillations Quantization~\cite{nagel2022overcoming} addressed oscillation issues by introducing a regularization term that encourages latent weights to be close to the center of the bin. ReBNN\cite{xu2023resilient} introduced weighted reconstruction loss to establish an adaptive training objective. The balance parameter associated with the reconstruction loss controls the weight oscillations.

In addition to addressing quantization-specific issues, some approaches integrate quantization with other training paradigms. SSQL~\cite{cao2022synergistic} unified self-supervised learning and quantization to learn quantization-friendly visual representations during pre-training. ~\cite{9147065} combined unsupervised learning and binary quantization to improve network compression and maintain model performance. \cite{wei2018quantization} effectively combined quantization and distillation, inducing the training of lightweight networks with solid performance.

\textbf{Maximum Entropy for Quantization.} Q-ViT~\cite{li2022q} believes that if one wants to restore the feature representation of FP model, the mutual information between the quantized model and the FP model should be maximized, so the quantized model should have the maximum information entropy. This work aligns with the goal of our work, but they assume that the output is Gaussian distribution and directly optimize the maximum entropy of the Gaussian distribution. This is clearly suboptimal, as high-dimensional features are not Gaussian distributions and are closer to long-tailed distributions.

N2UQ~\cite{liu2022nonuniform} also believes that more in formation are preserved when quantified quantized contain higher entropy. It applies a regularization term to the weights and optimizes them to give the quantized weights greater information entropy. The optimization object of this work is quantized  weights, while our optimization object is the learned representation

\section{Methodology}

%This section introduces our approach, MEC for quantization. We begin with the basic notation and a brief review of previous works, followed by our algorithm and analysis.

\subsection{Notation and Background}

\textbf{Basic Notations.} In this paper, notation $\mathbf{X}$ represents a matrix (or tensor), the vector is denoted as  $\mathbf{x}$, labeled data is $\mathbf{x}_l$, and unlabeled data is $\mathbf{x}_u$. $S(\mathbf{x};\mathbf{w}$) represents the model  $S\left ( \cdot  ;\cdot \right )$ with the parameter $\mathbf{w}$ and the input $\mathbf{x}$.

For a neural network with activation, we denote the loss function as $L\left (\mathbf{w},\mathbf{x} \right )$, where $\mathbf{w}$ and $\mathbf{x}$ represent the network's weights and input, respectively. Note that we assume $\mathbf{x}$ is sampled from the training set $\mathcal{D}_{t}$, thus the final loss is defined as $\mathbb{E}_{\mathbf{x}\sim \mathcal{D}_{t}} [L\left (\mathbf{w},\mathbf{x} \right)]$.

\textbf{Quantization.} Quantization parameters steps $s$ and zero points $z$ serve as a bridge between floating-point and fixed-point representations. Given the input tensor $\mathbf{x}$, the quantization operation is as follows:
\begin{equation}\label{eqt:quantization}
    \begin{aligned}
    \mathbf{x}_{int} &= clip\left (   \lfloor{\frac{\mathbf{x}}{s}}\rceil + z,0,2^{q}-1 \right )\\
    \hat{\mathbf{x}} &=\left ( \mathbf{x}_{int}-\mathbf{z}  \right ) s,
    \end{aligned}
\end{equation}

where $\lfloor{\cdot  }\rceil$ represents the rounding-to-nearest operator, $q$ is the predefined quantization bit-width, $\mathbf{s}$ denotes the step size between two subsequent quantization levels.  $\mathbf{z}$ stands for the zero-points. The $s$ and $z$ is initialized through forward propagation by a calibration set $\mathcal{D}_{c}$ ($\mathcal{D}_{c}\in \mathcal{D}_{t}$) from the training dataset $ \mathcal{D}_{t}$.
\begin{equation}\label{eq2}
s =\frac{\mathbf{x}_{max} - \mathbf{x}_{min}}{2^{q} - 1}.
\end{equation}
Different from LSQ~\cite{esser2019learned}, the quantization parameter $\mathbf{s}$ is calibrated by samples. Therefore, the loss function of a quantized model is given as follows:
\begin{equation}\label{eq3}
 \mathop{\arg\min}_{\hat{\mathbf{w}}} \ \  \mathbb{E}_{\mathbf{x}\sim \mathcal{D}_{t}} [L\left (\hat{\mathbf{w}},\mathbf{x} \right)],
\end{equation}
where $\hat{\mathbf{w}}$ is the quantized weight obtained from the latent weight $\mathbf{w}$ after quantization by~\eqref{eqt:quantization}.

\subsection{Feature Collapse for low bit quantization}
\label{sec:fea_col}
%https://arxiv.org/pdf/2310.04400

Singular value decomposition has been widely used to measure the collapse phenomenon (Jing et al., 2021). In Figure xxx, we have shown that the learned feature matrices $Z$ of the last layer are approximately low-rank with ???? some extremely small singular values. To determine the
degree of collapse for such matrices with low-rank tendencies, we propose the rectified eigenvalue entropy as a generalized
quantification as follows:
\begin{equation}
\label{eqt:fea_col}
H(x)=-\frac{rank(Z)}{N_{rank}}\sum_{i}\frac{\sigma_i}{\|\sigma_i\|}\log \frac{\sigma_i}{\|\sigma_i\|}
\end{equation}
the feature matrix should be diverse~\cite{} (full rank) and representative~\cite{} (the max eignvalue / min eignvalue = 1).

We conduct experiments on the CIFAR-10 dataset with the Quant-mec method, LSQ method, and full-precision ResNet-18 model. We randomly use 500 images from the CIFAR-10 dataset as test samplesWe use \eqref{eqt:fea_col} to measure the numerical values of feature collapse for Quant-MEC, LSQ, and full-precision models. The experimental results are shown in Tab. \ref{tab:fea_col}. The results indicate that the eigenvalues of features in the full-precision model are more evenly distributed, implying better feature representation. The eigenvalue distribution of the LSQ model is more concentrated, meaning only a few eigenvalues play a more critical role, facing the phenomenon of feature collapse. However, Quant-MEC alleviates this phenomenon.

\begin{table}[h!]
  \centering
  \caption{Numerical Comparison of feature collapse among Quant-MEC, LSQ and Full-Precision model.}
  \label{tab:fea_col}
\begin{tabular}{ccc}
\toprule
 & Quant-MEC & LSQ \\
 \midrule
Full Prec. & \textbf{8.598} $+-$ 0.000406 &  \\
\midrule
W4A4 & 7.812$+-$0.000008 & 6.220$+-$0.000067 \\
W2A4 & 7.163$+-$0.000089 & 6.404$+-$0.000139 \\
W2A2 & 7.804$+-$0.000124 & 6.497$+-$	0.000174 \\

\bottomrule
\end{tabular}
\end{table}

Intuitively, a matrix with high information abundance demonstrates a balanced distribution in vector space since
it has similar singular values. In contrast, a matrix with low information abundance suggests that the components
corresponding to smaller singular values can be compressed without significantly impacting the result. yet it
is applicable for non-strictly low-rank matrices, especially for fields with $D_i>> K$ which is possibly of rank K. We
calculate the information abundance of embedding matrices
for the enlarged DCNv2 (Wang et al., 2021) and compare
it with that of randomly initialized matrices, shown in Figure 2. It is observed that the information abundance of
learned embedding matrices is extremely low, indicating the
embedding collapse phenomenon.

\subsection{Maximum Entropy Coding}
\label{sec:MEC}

The framework of MEC-Quant is shown in Fig.\ref{fig:qat-kd-mec-comparision}(c). The model $S\left (\mathbf{x} ; \mathbf{w}\right )$ is the ones that we aim to quantize. It is first initialized with the calibration dataset.

In information theory, entropy for continuous variables is usually defined as: $H\left ( \mathbf{z} \right ) = -\int p\left ( \mathbf{z} \right ) \log_{}{p\left ( \mathbf{z} \right ) } dz$. In deep learning, the learned representation are usually high-dimensional vectors: $\mathbf{Z}=\left [\mathbf{z_{1}}, \mathbf{z_{2}},\dots  ,\mathbf{z_{m}}  \right ]$. However, it is very difﬁcult to estimate the true distributions~\cite{beirlant1997nonparametric, paninski2003estimation} $p(\mathbf{Z})$ from the learned representation $\mathbf{Z}$. A handy fact is that entropy is conceptually equivalent to the minimal number of bits required to encode the data losslessly, so the minimal lossless coding length could be used to represent the entropy. However, lossless coding of continuous random variables is infeasible in our case since it often requires an inﬁnite number of bits, breaking the numerical stability. Instead, we exploit the coding length in lossy data coding~\cite{cover1999elements} as a computationally tractable surrogate for the entropy of continuous random variables. For the learned representation $Z$, the minimal number of bits needed to encode Z subject to a distortion $\epsilon  $ is given by the following coding length function:

\begin{equation}\label{eq:mec_ori}
L= \left ( \frac{m+d}{2}  \right ) \mathrm{log} \;\mathrm{det}\left ( \mathbf{I}_{m
} +\frac{d}{m\epsilon ^{2} } \mathbf{Z^{\top}}  \mathbf{Z} \right ),
\end{equation}
where $\mathbf{I}_{m}$ denotes the identity matrix with dimension m, and $\epsilon$ is the upper bound of the expected decoding error between $\mathbf{z}\in \mathbf{Z}$ and the decoded $\mathbf{\hat{z}} $, i.e., $\mathbb{E}\left [ \left \| \textbf{z}- \hat{\textbf{z}}  \right \|  \right ] _{2} \le \epsilon $.

Due to the high computational cost of the log-determinant of the high-dimensional matrix in~\eqref{eq:mec_ori} and the potential for numerical instability of the ill conditioned matrix, we need to rewrite it. According to~\cite{liu2022self}, ~\eqref{eq:mec_ori} can be rewrite by Taylor expansion as:
\begin{equation}\label{eq:mec_tay}
L= \mathrm{Tr} \left ( \mu\sum_{k=1}^{\infty } \frac{\left (-1  \right )^{k+1}  }{k}   \left ( \lambda \textbf{Z}^{\top}\textbf{Z}  \right )^{k}   \right ) ,
\end{equation}
where $\mu =\frac{m+d}{2} $ and $\lambda =\frac{d}{m\epsilon ^{2} } $, with convergence condition: $\left \| \lambda \textbf{Z}^{\top}  \textbf{Z} \right \| _{2} < 1$.

\subsection{MoE and Gating}
\label{sec:MOE}

We believe that in deep learning, features typically exhibit a long-tailed distribution, and for~\eqref{eq:mec_tay}, it only unfolds at zero and cannot satisfy convergence conditions. Therefore, in order to address the problems faced by long-tailed distributions, we propose using the MoE mechanism to perform multi-point expansion on~\eqref{eq:mec_tay}.

In our work, each expert corresponds to the Taylor expansion of~\eqref{eq:mec_tay} at a certain point, and we assume that $f\left( \textbf{X} \right) = \log \left( \textbf{I}_{m} + \textbf{X} \right)$, where $\textbf{X} = \frac{d}{{m{\varepsilon ^2}}}{\textbf{Z}^ \top }\textbf{Z}$, $\textbf{X} \in \mathbb{R}^{m\times m}$, $\textbf{A}$ is the unit matrix with dimension $m\times m$. The n-th derivative of $f\left ( \textbf{X} \right ) $ is: ${f^{\left( n \right)}}\left( \textbf{X} \right) = {\left( { - 1} \right)^{n + 1}}\frac{{\left( {n - 1} \right)}}{{{{\left( {1 + {\rm{x}}} \right)}^n}}}$. The Taylor expansion of $f\left ( \textbf{X} \right )$ at  $\textbf{X}=a\cdot \textbf{A}$ is:
\begin{equation}\label{eq:multi_tay}
f(\textbf{X}) = f(a \cdot \textbf{A}) + \sum\limits_{k = 1} {\frac{{{f^{(k)}}\left( a \right)}}{{k!}}{{\left( {\textbf{X} - a \cdot \textbf{A}} \right)}^k}}.
\end{equation}
Therefore,~\eqref{eq:mec_tay} can be rewritten as:
\begin{equation}\label{eq:mec_multi_tay}
{L_i}\left( {\textbf{X};{a_i,k}} \right) = \mathrm{Tr}\left( {\mu \left( {f({a_i} \cdot A) + \sum\limits_{k = 1} {\frac{{{f^{(k)}}\left( {{a_i}} \right)}}{{k!}}{{\left( {X - {a_i} \cdot A} \right)}^k}} } \right)} \right),
\end{equation}
where $L_{i}$ represents the loss calculated by the i-th expert, and $\textbf{X}=a\cdot \textbf{A}$ represents the expansion point responsible by the i-th expert.

How to allocate the corresponding weights for each expert. We introduce the Gating Network mechanism, which is responsible for processing the weight coefficients of each expert.

\begin{equation}\label{eqt:moe}
    \begin{aligned}
    {L_{MEC}}\left( \textbf{Z} \right) = \sum\limits_{i = 1}^n {\left( {G{{\left( \textbf{Z} \right)}_i}{L_i}\left( {\textbf{X};{a_i,k}} \right)} \right)} \\
    G\left( \textbf{Z} \right) = \mathrm{Softmax}\left( {{W_g}\left(\textbf{Z} \right)} \right),
    \end{aligned}
\end{equation}
where $G\left ( \cdot  \right ) $ represents the gating network, $W_{g}$ is the corresponding parameter of the gated network, $n$ represents the number of experts in MoE, $\textbf{X} = \frac{d}{{m{\varepsilon ^2}}}{\textbf{Z}^ \top }\textbf{Z}$. After Softmax processing, $\textbf{Z}$ is assigned weights to each expert. The final calculation result of MoE is the weighted sum of $n$ expert network outputs.

\subsection{Training Processing of MEC-Quant}\label{sec:training}
In this paper, we focus on the classification task. Therefore, the whole loss consists of two components as follows:
\begin{equation}\label{eqt:compositedloss}
{L_{toal}}\left( {\hat{\textbf{w}} ,\textbf{x}} \right) = task\_loss\left (  \textbf{Z}_o\right ) + \lambda {L_{MEC}}\left( \textbf{Z}_{b} \right),
\end{equation}
where $\mathbf{Z}_{o}$ presents the output from the model for the labeled data $(\mathbf{x},y)$, $\mathbf{Z}_{b}$ presents the output from the backbone of model, weight $\lambda$ ($\lambda>0$) balances between the CE loss and MEC loss in~\eqref{eqt:compositedloss}.

In this work, we focus on classification tasks, and there are two possible case for task loss:
\begin{equation}\label{eqt:task_loss_ce}
task\_loss({\textbf{Z}_o}) = CE({\textbf{Z}_o},y),
\end{equation}
where $CE\left ( \cdot ,\cdot  \right )$ denotes the cross-entropy loss for classification tasks, $y$ represents the annotation information of the data. In this case, we use the annotations of the data as supervised information. Another case is:
\begin{equation}\label{eqt:task_loss_kl}
task\_loss({\textbf{Z}_o}) = KL({\textbf{Z}_o},{\textbf{Z}_t}),
\end{equation}
where  $KL(\cdot ,\cdot )$ represents the calculation of KL divergence, $\textbf{Z}_t$ is the output of the FP model. In this case, we use the features of the full precision model as supervised information. In this case, MEC-Quant is a label free quantification.

In order to avoid trivial solutions, the model should have a certain prior information as a guarantee that the model learns in the direction related to downstream tasks, or in other words, the model has its own main driving force. In~\eqref{eqt:task_loss_ce}, the annotation information $y$ is our prior information, and annotation $y$ ensures that the quantization model learns in the correct direction for classification. In~\eqref{eqt:task_loss_kl}, the feature representation of the FP model is our prior information, and the fully accurate feature representation ensures that the quantized model learns from the FP feature space. Therefore, MEC constraints can be seen as an auxiliary regularization.

\begin{figure}[t!]
    \centering
    \includegraphics[width=0.85\linewidth]{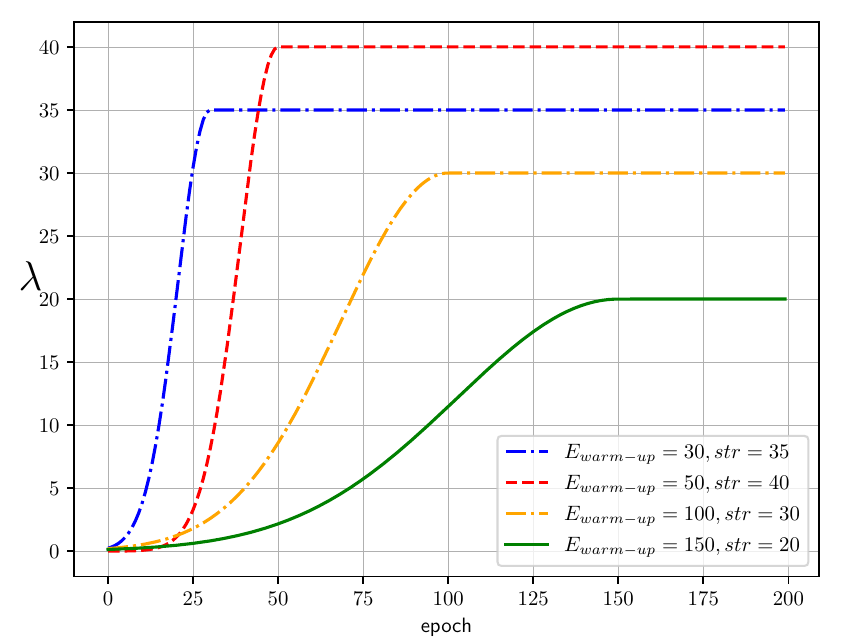}
    \caption{Different settings of the strength ($\lambda$) of MEC}
    \label{fig:lambda}
\end{figure}

The weight $\lambda$  is very important in ~\eqref{eqt:compositedloss}. We hope that in the early stage of QAT training, the strength of MEC constraints should be as small as possible, with the aim of allowing the model to fully learn the priors related to downstream tasks. In the later stage of QAT training, the strength of MEC constraints should increase in order to enable the model to have greater feature entropy while maintaining performance. In summary, the intensity of MEC constraints should gradually increase with the training rounds, reaching its maximum intensity in the middle stage and remaining constant in the later stages of training. In this paper, $\lambda$ is progressively increased as follows:
\begin{equation}\label{eqt:lambda}
 \lambda =\text{str}\times e^{-5\times \left [ 1-\left ( \frac{\beta}{E_{\text{warm-up}}}  \right )^{2}   \right ] },
\end{equation}
where $\text{str}$ represents the intensity of MEC, $E_{\text{warm-up}}$ represents the predetermined hyperparameter of warm-up epoch, and $\beta$ is:
\begin{equation}\label{eqt:betainlambda}
 \beta= clip\left ( t,0,E_{\text{warm-up}} \right ),
\end{equation}
where $t$ represents the current epoch.

As illustrated in Fig.~\ref{fig:lambda}, a small $\lambda$ value in~\eqref{eqt:lambda} can ensure that the model learn from the ground truth in the early stages of training, in order to prevent the MEC constraints in the later stages from obtaining the trivial solution. The $\lambda$ value gradually increases, and while ensuring that task loss is dominant, the entropy of the features gradually increases. Ultimately, a stable $\lambda$ ensures that MEC constraints play a role correctly. Our overall training algorithm is shown in algorithm~\ref{alg:mec}. The overall algorithm is very simple and effective.

\begin{algorithm}[t!]
   \caption{MEC-Quant}
   \label{alg:mec}
\begin{algorithmic}[1]
   \STATE {\bfseries Input:} 
      labeled data $\mathbf{x}$;
      quantized model $S\left ( \mathbf{x} ; \mathbf{w}\right )$;
      iteration $T$.
   \FOR{$i=1$ {\bfseries to} $T$}
        \STATE Preprocess the input data;
        \STATE Feed the data into the quantized model, get the final output of the model $\textbf{Z}_{o}$ and the output of the backbone $\textbf{Z}_{b}$;
        \STATE Calculate the MEC Loss and the task loss based on~\eqref{eqt:moe} and~\eqref{eqt:compositedloss};
        \STATE Update for student model's weights parameter and quantization parameter;
    \ENDFOR \\
    \STATE {\bfseries Output:} 
      quantized model.
\end{algorithmic}
\end{algorithm}

\section{Experiments}

%In this section, we conduct three sets of experiments to verify the effectiveness of MEC-Quant. In Sec.\ref{sec4.1}, we first conduct ablation experiments to investigate the impact of MEC regularization and MoE mechanism on CIFAR-10~\cite{krizhevsky2009learning} dataset. In Sec.\ref{sec:parameter-configuration}, we experimentally analyze and discuss our hyperparameter settings. In Sec.\ref{sec4.3}, we compare our method with other effective QAT methods on CIFAR-10 and CIFAR-100~\cite{krizhevsky2009learning} datasets.

\textbf{Experimental Protocols.} Our code is based on PyTorch~\cite{paszke2019pytorch} and relies on the MQBench~\cite{li2021mqbench} package. By default, we set the $\beta$ increase linearly from 0 to 5 as the number of epochs increases unless explicitly mentioned otherwise. The default data augmentation includes random horizontal flipping and random translation. As for quantization, we use the default method of asymmetric quantization. For the cifar-10 and cifar-100 datasets, we use 100 images for calibration. We also keep the first and last layers with 8-bit quantization, which is same as QDrop~\cite{wei2022qdrop}. Additionally, we employ per-channel quantization for weight quantization. We use WXAX to represent X-bit weight and activation quantization. 

Two experimental settings are evaluated as follows:
\begin{itemize}
\item[] (A) We use annotation information $y$ as supervised information. At this time, the annotation information $y$ is the prior information of MEC-Quant, as illustrated in~\eqref{eqt:task_loss_ce}.
\item[] (B) We use the output feature $\textbf{Z}_t$ of the FP model as supervised information. At this time, the output feature $\textbf{Z}_t$ of the FP model is the prior information of MEC-Quant, as illustrated in~\eqref{eqt:task_loss_kl}.
\end{itemize}

\textbf{Training Details.} We use SGD as the optimizer, with a batch size of 256 and a base learning rate of 0.01. The default learning rate (LR) scheduler follows the cosine annealing method. The weight decay is 0.0005, and the SGD momentum is 0.9. We train for 200 epochs on CIFAR-10 and CIFAR-100 unless otherwise specified.

\subsection{Ablation Study}
\label{sec4.1}

\begin{table}[h!]
\setlength{\belowcaptionskip}{0.3cm}
\begin{center}
\centering
\caption{Ablation studies of the choice of MEC-Qaunt (Accuracy $\%$) on CIFAR-10.}
\label{tbl:ablation}
\resizebox{.4\textwidth}{!}{
\begin{tabular}{cccc}
\toprule
\begin{tabular}[c]{@{}l@{}}
MEC
\end{tabular} &
\begin{tabular}[c]{@{}l@{}} MoE
\end{tabular} &
\begin{tabular}[c]{@{}l@{}}ResNet-18
\end{tabular} &
\begin{tabular}[c]{@{}l@{}}MobileNetV2
\end{tabular}

\\ \midrule
                  &                               &88.36                                 &78.15                             \\
 \checkmark       &                               &88.90                                     &   81.03                          \\
 \checkmark      & \checkmark                    &\textbf{88.98}                         &\textbf{81.20}                             \\
\bottomrule
\end{tabular}
}
\end{center}
\end{table}

\textbf{Effectiveness of MEC:} To verify the effectiveness of MEC and MoE, we used W2A4 quantization configuration on CIFAR-10 dataset. In Tab.~\ref{tbl:ablation}, we used LSQ as our baseline, with ResNet-18~\cite{he2016deep} having an accuracy of 88.36$\%$ and MobileNetV2~\cite{sandler2018mobilenetv2} having an accuracy of 78.15$\%$. When using MEC validation experiments, we use~\eqref{eq:mec_tay} as our regularization constraint and use 4th order Taylor expansion. When using MoE in combination, we reduced the Taylor expansion of MEC~\eqref{eqt:moe} to 2th order to reduce computational complexity and address issues arising from long-tailed distributions.

\subsection{Parameter Configuration}
\label{sec:parameter-configuration}

In this section, We investigate the effect of hyperparameters on MEC-Quant experiments.

\begin{table}[h!]
\setlength{\belowcaptionskip}{0.3cm}
\begin{center}
\centering
\caption{Hyperparameter comparative experiment of the strength of MEC on CIFAR-10.}
\label{tbl:strength}
\begin{tabular}{lc}
\toprule
\multicolumn{1}{c}{The strength of MEC.}                                            & \begin{tabular}[c]{@{}c@{}} Val Acc(\%)\end{tabular}       \\ \midrule
\begin{tabular}[c]{@{}l@{}}Constant 1 (baseline)\end{tabular}                                  & 87.85                                     \\\hline\hline
\begin{tabular}[c]{@{}l@{}}\eqref{eqt:lambda}, where \text{str=0.5}, $E_{\text{warm-up}}=50$\end{tabular}                 & 88.07 \\
\begin{tabular}[c]{@{}l@{}}\eqref{eqt:lambda}, where \text{str=5}, $E_{\text{warm-up}}=50$ \end{tabular}                   & \textbf{88.26}  \\
\begin{tabular}[c]{@{}l@{}}\eqref{eqt:lambda}, where \text{str=10}, $E_{\text{warm-up}}=50$\end{tabular}                   & 88.17 \\
\begin{tabular}[c]{@{}l@{}}\eqref{eqt:lambda}, where \text{str=5}, $E_{\text{warm-up}}=100$\end{tabular}                   & 88.02 \\
\begin{tabular}[c]{@{}l@{}}\eqref{eqt:lambda}, where \text{str=5}, $E_{\text{warm-up}}=30$\end{tabular}                   & 88.17 \\ \bottomrule
\end{tabular}
\end{center}
\end{table}

\textbf{The Strength of MEC.} MEC constraint strength builds a bridge between task loss and MEC constraints. We quantized the ResNet18 model using W2A4 on the CIFAR-10 dataset, and the results are shown in Tab.~\ref{tbl:strength}. As mentioned in section~\ref{sec:training}, MEC constraints can be seen as an auxiliary regularization. In the early stage of QAT training, the strength of MEC constraints should be as small as possible, with the aim of allowing the model to fully learn downstream task related priors. In the later stage of QAT training, the strength of MEC constraints should increase in order to enable the model to have greater feature entropy while maintaining performance.

\begin{table}[h!]
\setlength{\belowcaptionskip}{0.3cm}
\begin{center}
\centering
\caption{Hyperparameter comparative experiment of Taylor expansion method on CIFAR-10.}
\label{tbl:taylor_order}
\begin{tabular}{lc}
\toprule
\multicolumn{1}{c}{Taylor expansion method.}                                            & \begin{tabular}[c]{@{}c@{}} Val Acc(\%)\end{tabular}       \\ \midrule
\begin{tabular}[c]{@{}l@{}}\eqref{eq:mec_tay}, where $k=4$\end{tabular}                 & 89.85 \\
\begin{tabular}[c]{@{}l@{}}\eqref{eq:mec_tay}, where $k=2$ \end{tabular}                   & 89.67  \\
\begin{tabular}[c]{@{}l@{}}\eqref{eqt:moe}, where $k=2$\end{tabular}                   & 89.98 \\ \bottomrule
\end{tabular}
\end{center}
\end{table}

\textbf{Taylor expansion order.} We quantized the ResNet18 model using W2A2 on the CIFAR-10 dataset, and the results are shown in Tab.~\ref{tbl:taylor_order}. When using~\eqref{eq:mec_tay}, as the order of Taylor expansion increases, the model performance improves, but at the same time, it is accompanied by a larger computational load. When using~\eqref{eqt:moe}, we found that using only the Taylor expansion with k=2 can achieve the effect of k=4 in~\eqref{eq:mec_tay}.

\subsection{Literature Comparison}
\label{sec4.3}

\begin{table}[h!]
\begin{center}
\centering
\caption{Comparison among different QAT strategies regarding accuracy on CIFAR-10.}
\label{tbl:comp-CIFAR-10}
\resizebox{0.85\linewidth}{!}{
\begin{tabular}{clrcccc}
\toprule
\multicolumn{1}{c}{\begin{tabular}[c]{@{}c@{}}\textbf{Labeled} \\ \textbf{data}\end{tabular}} & \multicolumn{1}{l}{\textbf{Methods}} & \multicolumn{1}{c}{\textbf{W/A}} & \multicolumn{1}{c}{\textbf{Res18}} & \multicolumn{1}{c}{\textbf{Res50}}  & \multicolumn{1}{c}{\textbf{MBV1}} & \multicolumn{1}{c}{\textbf{MBV2}} \\ \midrule
50000                       & Full Prec.                           & 32/32 & 88.72         &89.95                     &85.52           &85.81      \\ \hline\hline
\multirow{18}{*}{50000}     & PACT~\cite{choi2018pact}             & 4/4   &88.15          &85.27                     &80.77           &79.88      \\
                            & LSQ~\cite{esser2019learned}          & 4/4   &88.69          &90.01                     &82.39           &84.45      \\
                            & LSQ+~\cite{bhalgat2020lsq+}          & 4/4   &88.40          &90.30                     &84.32           &84.30  \\
                            & KD~\cite{hinton2015distilling}       & 4/4   &88.86          &90.34                     &84.77           &83.79  \\
                            & MEC-Qaunt (Setting (A))              & 4/4   &\textbf{90.22} &\textbf{91.25}            &\textbf{85.44}  &\textbf{85.52}    \\
                            & MEC-Qaunt (Setting (B))              & 4/4   &90.18          &90.88                     &85.32           &\textbf{85.57}     \\ \cmidrule{2-7}
                            & PACT~\cite{choi2018pact}             & 2/4   &87.55          &85.24                      &69.04           &67.18      \\
                            & LSQ~\cite{esser2019learned}          & 2/4   &88.36          &90.01                      &78.15           &78.15      \\
                            & LSQ+~\cite{bhalgat2020lsq+}          & 2/4   &87.76          &89.62                      &81.26           &77.00      \\
                            & KD~\cite{hinton2015distilling}       & 2/4   &88.83          &90.18                      &78.84           &75.56  \\
                            & MEC-Qaunt (Setting (A))              & 2/4   &89.76          &91.11                      &\textbf{80.79}  &\textbf{81.20}      \\  
                            & MEC-Qaunt (Setting (B))              & 2/4   &\textbf{89.98} &\textbf{91.23}             &80.55           &80.96      \\ \cmidrule{2-7}
                            & PACT~\cite{choi2018pact}             & 2/2   &76.90          &64.94              &11.71           &10.58        \\
                            & LSQ~\cite{esser2019learned}          & 2/2   &87.60          &87.79              &75.29          &70.32        \\
                            & LSQ+~\cite{bhalgat2020lsq+}          & 2/2   &87.60          &86.10              &74.22           &72.18         \\
                            & KD~\cite{hinton2015distilling}       & 2/2   &88.06          &89.20              &68.62           &67.15         \\
                            & MEC-Qaunt (Setting (A))              & 2/2   &\textbf{88.51} &89.15              &75.86           &72.87      \\
                            & MEC-Qaunt (Setting (B))              & 2/2   &88.26          &\textbf{89.32}     &\textbf{76.02}  &\textbf{73.77}      \\\bottomrule

\end{tabular}
}
\end{center}
\end{table}

\textbf{CIFAR-10.} We selected ResNet-18 and -50~\cite{he2016deep} with normal convolutions, MobileNetV1~\cite{howard2017mobilenets} and V2~\cite{sandler2018mobilenetv2} with depth-wise separable convolutions as our experimental models. In Tab.~\ref{tbl:comp-CIFAR-10}, we quantized the weights to 2-bit and the activations to 2-bit. We compared our approach with effective baselines, including LSQ~\cite{esser2019learned}, LSQ+~\cite{bhalgat2020lsq+}, PACT~\cite{choi2018pact}, and KD~\cite{hinton2015distilling}. 

As shown in Tab.~\ref{tbl:comp-CIFAR-10}, in setting (A), MEC-Quant showed significant improvements compared to the baselines. In W4A4 quantization, MEC-Quant achieved about 1$\sim$2$\%$ accuracy improvements over LSQ. In W2A4 quantization, MEC-Quant achieved nearly 3$\%$ accuracy improvements over LSQ of MobilenetV2 model. Furthermore, to explore the ability of MEC-Qaunt, we conducted W2A2 quantization experiment. In W2A2 quantization, the advantages of MEC-Quant are more obvious. It achieved 3.4$\%$ accuracy improvements over LSQ of MobilenetV2 model. A similar situation occurred in setting (B). The accuracy of setting (B) is very close to that of setting (A).

\begin{table}[h!]
\begin{center}
\centering
\caption{Comparison among different QAT strategies regarding accuracy on CIFAR-100.}
\label{tbl:comp-CIFAR-100}
\resizebox{0.85\linewidth}{!}{
\begin{tabular}{clrcccc}
\toprule
\multicolumn{1}{c}{\begin{tabular}[c]{@{}c@{}}\textbf{Labeled} \\ \textbf{data}\end{tabular}} & \multicolumn{1}{l}{\textbf{Methods}} & \multicolumn{1}{c}{\textbf{W/A}} & \multicolumn{1}{c}{\textbf{Res18}} & \multicolumn{1}{c}{\textbf{Res50}}  & \multicolumn{1}{c}{\textbf{MBV1}} & \multicolumn{1}{c}{\textbf{MBV2}} \\ \midrule
50000                       & Full Prec.                           & 32/32 &75.40          & 78.94                    &70.22           &71.30      \\ \hline\hline
\multirow{18}{*}{50000}     & PACT~\cite{choi2018pact}             & 4/4   &74.17          &74.78                     &64.65           &64.06      \\
                            & LSQ~\cite{esser2019learned}          & 4/4   &75.30           &78.20                    &68.63           &69.01      \\
                            & LSQ+~\cite{bhalgat2020lsq+}          & 4/4   &74.50           &77.39                    &67.89           &68.25  \\
                            & KD~\cite{hinton2015distilling}       & 4/4   &74.70           &78.80                    &\textbf{70.96}  &\textbf{71.66} \\
                            & MEC-Qaunt (Setting (A))              & 4/4   &\textbf{75.65} &\textbf{78.65}            &68.75           &68.51    \\ 
                            & MEC-Qaunt (Setting (B))              & 4/4   &\textbf{75.98} &\textbf{79.50}            &70.11           &70.33     \\ \cmidrule{2-7}
                            & PACT~\cite{choi2018pact}             & 2/4   &73.77          &74.72                      &49.98           &57.90      \\
                            & LSQ~\cite{esser2019learned}          & 2/4   &74.93          &77.82                      &65.13           &66.15      \\
                            & LSQ+~\cite{bhalgat2020lsq+}          & 2/4   &73.90          &76.61                      &65.28           &\textbf{66.24}      \\
                            & KD~\cite{hinton2015distilling}       & 2/4   &74.35          &77.34                      &\textbf{66.90}  &63.77  \\
                            & MEC-Qaunt (Setting (A))              & 2/4   &\textbf{75.2}  &77.72                      &65.97           &65.00      \\ 
                            & MEC-Qaunt (Setting (B))              & 2/4   &\textbf{75.79} &\textbf{79.22}             &\textbf{66.75}  &64.61      \\ \cmidrule{2-7}
                            & PACT~\cite{choi2018pact}             & 2/2   &65.16          &4.26                      &3.25            &8.39        \\
                            & LSQ~\cite{esser2019learned}          & 2/2   &71.80          &62.79                     &\textbf{55.53}  &31.08        \\
                            & LSQ+~\cite{bhalgat2020lsq+}          & 2/2   &71.25          &84.21                     &55.56           &30.08          \\
                            & KD~\cite{hinton2015distilling}       & 2/2   &73.13          &63.16                     &55.37           &28.56         \\
                            & MEC-Qaunt (Setting (A))              & 2/2   &\textbf{73.57} &\textbf{64.8}     &\textbf{58.98}  &\textbf{34.54}      \\
                            & MEC-Qaunt (Setting (B))              & 2/2   &\textbf{75.15} &\textbf{71.54}     &\textbf{62.99}  &\textbf{42.54}      \\\bottomrule

\end{tabular}
}
\end{center}
\end{table}

\textbf{CIFAR-100} In this section, we investigated the effectiveness of MEC-Quant on the CIFAR-100 dataset. We employed ResNet-18, ResNet-50, MobileNetV1 and MobileNetV2 as the experimental models. In Tab.~\ref{tbl:comp-CIFAR-100}, we quantized the weights to 2-bit and the activations to 2-bit. Similar to CIFAR-10, We compared our approach with effective baselines, including LSQ~\cite{esser2019learned}, LSQ+~\cite{bhalgat2020lsq+},  PACT~\cite{choi2018pact}, and KD~\cite{hinton2015distilling}.%另外加一个ghostnet.

%\textbf{super resolution}

%\textbf{图像降噪或分割}

Similar to CIFAR-10, MEC-Quant showed significant improvements compared to the baselines. As shown in Tab6, In W4A4 quantization, MEC-Quant achieved about x$\sim$x$\%$ accuracy improvements over LSQ. In W2A4 quantization, MEC-Quant achieved nearly x$\sim$x$\%$ accuracy improvements over LSQ of MobilenetV2 model. Furthermore, to explore the ability of MEC-Quant, we conducted W2A2 quantization experiment. In W2A2 quantization,the advantages of MEC-Quant are more obvious. It achieved 3.4$\%$ accuracy improvements over LSQ of MobilenetV2 model.

\subsection{Generalization of MEC-Quant}
\label{sec:Analysis}

\begin{table}[h!]
  \centering
  \caption{A comparison of the Hessian eigenvalues of the models trained by the MEC-Quant and LSQ methods under W2A2.}
  \label{tab:hessian}
  \begin{tabular}{lcc}
    \toprule
    \textbf{method} & \textbf{max eigenvalue} & \textbf{mean eigenvalue} \\
    \midrule
    MEC-Quant & 155.73 & 18.7 \\
    LSQ       & 520.24 & 61.65 \\
    \bottomrule
  \end{tabular}
  \vspace{0.2cm}
\end{table}

It has been empirically pointed out that the dominant eigenvalue of $\nabla^2 L_{\text{val}}(\boldsymbol{w})$ (spectral norm of Hessian) is highly correlated with the generalization quality of QAT solutions~\cite{keskar-eigenvalue1-iclr-2017} \cite{Wen-eigenvalue2-icais-2020}. In standard QAT training, the Hessian norm is usually great, which leads to deteriorating (test) performance of the solutions. We calculated the eigenvalues of the Hessian matrices of MEC-Quant and the comparison method LSQ under W2A2, and obtained the results shown in Tab. \ref{tab:hessian}. The maximum eigenvalue of MEC-Quant is 155.73, which is much lower than 520.24 of LSQ. The average value of the eigenvalues of MEC-Quant is 18.7, which is also much lower than 61.65 of LSQ. Both phenomena indicate the advantage of MEC-Quant in enhancing generalization.

\section{Conclusion}

In this paper, we propose MEC-Quant, a simple, novel, yet compelling paradigm for QAT. MEC-Quant is a more principled objective that explicitly optimizes on the structure of the representation, so that the learned representation is less biased and thus generalizes better to unseen in-distribution samples. MEC-Quant pushes the performances of QAT models towards and surpasses that of FP32 counterparts, especially at the low bit widths. Our approach demonstrates promising results across various neural network models. Extensive experiments show that our method successfully enhances the generalization ability of the quantized model and outperforms the SOTA approaches in recent QAT research.

\bibliographystyle{IEEEtran}
\bibliography{example_paper}
\end{document}